\documentclass[9pt,review]{elsarticle}

\usepackage{hyperref}

\usepackage{amssymb,amsmath,bm,mathrsfs,amsfonts,graphicx,multirow,stmaryrd}

\usepackage{times}
\usepackage{natbib,dsfont}
\usepackage{amsthm}
\usepackage{color}
\usepackage{enumitem}
\usepackage{amssymb,amsmath,bm,mathrsfs,amsfonts,graphicx,multirow,stmaryrd,color}
\usepackage{amsmath,amssymb,amsthm,enumerate,epsfig,graphicx}
\usepackage{amsmath,amsthm,amssymb,amscd,txfonts}
\usepackage{times}
\usepackage{natbib,dsfont}
\usepackage{amsthm}
\usepackage{epstopdf}
\usepackage{graphics}
\usepackage{subfig}
\usepackage{float}
\usepackage{mathrsfs}
\usepackage{booktabs}
\usepackage{footnote}
\usepackage{graphicx}
\usepackage{epstopdf}
\usepackage{caption}
\usepackage{enumerate}

\numberwithin{equation}{section}
\theoremstyle{plain}
\theoremstyle{remark}

\newtheorem{theorem}{Theorem}

\newcommand{\bmu}{{\bm \mu}}
\newcommand{\bgamma}{{\bm \gamma}}
\newcommand{\bSig}{{\bm \Sigma}}

\newcommand{\bom}{{\bm \omega}}

\newcommand{\bD}{{\bf D}}

\newcommand{\bG}{{\bf G}}

\newcommand{\bu}{{\bf u}}

\newcommand{\bx}{{\bf x}}

\newcommand{\bqa}{\begin{eqnarray}}
\newcommand{\eqa}{\end{eqnarray}}
\newcommand{\bqn}{\begin{eqnarray*}}
	\newcommand{\eqn}{\end{eqnarray*}}
\newcommand{\be}{\begin{equation}}
\newcommand{\ee}{\end{equation}}
\allowdisplaybreaks[4]
\numberwithin{equation}{section}

\renewcommand{\(}{\left(}
\renewcommand{\)}{\right)}

\def\om{\omega}

\def\phi{\varphi}

\newtheorem{thm}{Theorem}[section]

\newtheorem*{prop}{Proof}

\theoremstyle{definition}

\numberwithin{equation}{section}
\numberwithin{thm}{section}

\theoremstyle{remark}
\newtheorem{assumption}[thm]{Assumption}

\def\fx {{\mathbf x}}

\def\tfH {{\widetilde{\mathbf H}}}

\def\fv {{\mathbf v}}
\def\fu {{\mathbf u}}
\def\fw {{\mathbf w}}
\def\fW {{\mathbf W}}
\def\fS {{\mathbf S}}

\def\fC {{\mathbf C}}

\def\fI {{\mathbf I}}

\def\bbI {{\mathbb I}}
\def\bbA {{\mathbb A}}
\def\bbB {{\mathbb B}}

\def\tfS {{\widetilde{\mathbf S}}}

\newcommand{\beq}{\begin{eqnarray}}
\newcommand{\eeq}{\end{eqnarray}}
\newcommand{\beqq}{\begin{eqnarray*}}
\newcommand{\eeqq}{\end{eqnarray*}}

\journal{Journal of \LaTeX\ Templates}

\begin{document}
	
\begin{frontmatter}	
	
\title{Spectrally-Corrected and Regularized LDA for Spiked Model}	

\author[mymainaddress]{Hua Li\fnref{myfootnote1}}

\author[mysecondaddress1]{Wenya Luo\corref{mycorrespondingauthor}}
\ead{luowy042@zufe.edu.cn}

\author[mysecondaddress2]{Zhidong Bai}
\author[mysecondaddress2]{Huanchao Zhou}
\author[mysecondaddress2]{Zhangni Pu}


\address[mymainaddress]{School of Science, Chang Chun University, China}
\address[mysecondaddress1]{School of Data Sciences, Zhejiang University of Finance and Economics, China}

\address[mysecondaddress2]{KLAS MOE and School of Mathematics and Statistics, Northeast Normal University, China}
	
\begin{abstract}
This paper proposes an improved linear discriminant analysis called spectrally-corrected and regularized LDA (SRLDA). This method integrates the design ideas of the sample spectrally-corrected covariance matrix and the regularized discriminant analysis. With the support of a large-dimensional random matrix analysis framework, it is proved that SRLDA has a linear classification optimal global optimal solution under the spiked model assumption. According to simulation data analysis, the SRLDA classifier performs better than RLDA and ILDA and is closer to the theoretical classifier. Experiments on different data sets show that the SRLDA algorithm performs better in classification and dimensionality reduction than currently used tools.
\end{abstract}	

\end{frontmatter}

\section{Introduction}

{L}{inear} Discriminant Analysis (LDA) is a very common technique for dimensionality reduction problems as a preprocessing step for machine learning and pattern classification applications. As data collection and storage capacity improves, there has been an increasing prevalence of high-dimensional data sets, including microarray gene expression data \cite{Pomeroy2002Prediction}, gene expression pattern images \cite{2004Identifying}, text documents \cite{2003Lower}, face images \cite{swets1996using}, etc. Learning in high-dimensional spaces is challenging because data points are far from each other in such spaces, and the similarities between data points are difficult to compare and analyze \cite{2000Aide}. This phenomenon is traditionally known as the curse of dimensionality \cite{1962Adaptive}, which states that an enormous number of samples is required to perform accurate predictions on problems with high dimensionality.

On the other hand, large-dimensional random matrix theory (LRMT) analysis tools have gradually developed and matured, and their applications in fields such as statistics, economics, computers, signal processing, and so on, have gained increasing recognition \cite{2021Large, couillet2022Liao, bai2014spectral}. In LRMT, researchers study the asymptotic properties of random matrices in the high-dimensional range, where the dimensions of the random matrices are extremely large or even infinite. The extreme results obtained in the infinite-dimensional case can well approximate the more realistic finite-dimensional scenario, and this has been verified by many empirical results. LRMT has proven to be a valuable tool for analyzing high-dimensional systems in a variety of applications, including signal processing, data analysis, and machine learning. In this paper, we provide an improved linear discriminant analysis for spiked models with LRMT.

The statistics problem treated here is assigning a $p$-dimensional observation $\fx=\(x_1,x_2,\dots,x_p\)'$ into one of two classes or groups $\Pi_i$ $(i=0,1)$. The classes are assumed to have Gaussian distributions with the same covariance matrix. Since R.A. Fisher \cite{fisher1936use} originally proposed Linear Discriminant Analysis (LDA) based classifiers in 1936, Fisher's LDA has been one of the most classical techniques for classification tasks and is still used routinely in applications. 
We employ separate (stratified) sampling: $n = n_0 + n_1$ sample points are collected to constitute the sample $\fC$ in $R^p$, where, given $n$, $n_0$ and $n_1$ are determined (not random) and where $\fC_0=\{{\bf x}_1, {\bf x}_2,...,{\bf x}_{n_0}\}$ and $\fC_1=\{{\bf x}_{n_0+1}, {\bf x}_{n_0+2},...,{\bf x}_n\}$ are randomly selected from populations $\Pi_0$ and $\Pi_1$, respectively. 
Under the support of the labeled sample sets, Fisher's discriminant rule \cite{johnson2002applied} with two multivariate normal populations directs us to allocate $\fx$ into $\Pi_0$ if
\beq\label{eq:wlda}
W^{LDA}(\overline{\fx}_0, \overline{\fx}_1, \fS, \fx)=\left(\fx-\frac{\overline{\fx}_0 + \overline{\fx}_1}{2}\right)^T\fS^{-1}
\left(\overline{\fx}_0-\overline{\fx}_1\right)
-\log\frac{\pi_1}{\pi_0}>0,
\eeq
and into $\Pi_1$ otherwise, where $\pi_i$ is the prior probability corresponding to $\Pi_i$ ($i=0,1$). Here $\overline{\fx}_0$ and $\overline{\fx}_1$ are the sample mean vectors of two classes, and $\fS = \frac{n_0}{n}\fS_0 + \frac{n_1}{n}\fS_1$ in which $\fS_i$ is the sample covariance matrix of $\fC_i$. LDA has a long and successful history. From the first use of taxonomic classification by \cite{fisher1936use}, LDA-Fisher-based classification and recognition systems have been used in many applications, including detection \cite{varshney2012generalization}, speech recognition \cite{vuuren1997data}, cancer genomics \cite{kim2002identification}, and face recognition \cite{swets1996using}.

In the classic asymptotic approach, $W^{LDA}(\overline{\fx}_0, \overline{\fx}_1, \fS, \fx)$ is the consistent estimation of Fisher's LDA function. However, it is not generally helpful in situations where the dimensionality of the observations is of the same order of magnitude as the sample size $(n\rightarrow\infty, p\rightarrow\infty,$ and $p/n\rightarrow J>0)$.
In this asymptotic scenario, the Fisher discriminant rule performs poorly due to the sample covariance matrix diverging from the population's one severely. Papers developing different approaches to handling the high-dimensionality issue in estimating the covariance matrix can be divided into two schools. The first suggests building on the additional knowledge in the estimation process, such as sparseness, graph model, or factor model \cite{bickel2008regularized, cai2012minimax,fan2008high,khare2011wishart,rajaratnam2008flexible, ravikumar2011high,rohde2011estimation}.
The second recommends correcting the spectrum of the sample covariance \cite{2021Spectrally}, such as the shrinkage estimator in \cite{el2008spectrum, ledoit2004well,ledoit2012nonlinear}, and regularized technologic in \cite{bakir2016efficient, elkhalil2017asymptotic, sifaou2020high, zollanvari2015generalized}. The Spectrally-Corrected and Regularized LDA  (SRLDA) given in this paper belong to the second school, which integrates the design ideas of the sample spectrally-corrected covariance matrix \cite{2021Spectrally} and the regularized discriminant analysis \cite{friedman1989regularized} to improve the LDA estimation in high-dimensional settings.

This paper proposes a novel approach under the assumption of the spiked covariance model in \cite{johnstone2001distribution} that all but a finite number of eigenvalues of the population covariance matrix are the same. This model could be and has been used in many real applications, such as detection speech recognition \cite{hastie1995penalized, johnstone2001distribution}, mathematical financial \cite{ 2008Determining, laloux2000random, malevergne210115collective,passemier2017estimation, plerou2002random}, wireless communication \cite{telatar1999capacity}, physics of mixture \cite{sear2003instabilities}, EEG signals \cite{2009Functional, fazli2011?1} and data analysis and statistical learning \cite{hoyle2003limiting}. Based on some theoretical and applied results of the spiked covariance model(\cite{bai2012estimation, 2006BaikSilverstein, 2022Bao, ke2021estimation, 2007Asymptotics}), we suppose population eigenvalues are estimated reasonably. Then consider a class of covariance matrix estimators that follow the same spiked structure, written as a finite rank perturbation of a scaled identity matrix. The sample eigenvectors of the extreme sample eigenvalues provide the directions of the low-rank perturbation, and the corresponding eigenvalues are corrected to population one and regularized by the common parameter. In this way, compared with \cite{sifaou2020high}, we not only retain the spike structure as much as possible, but also reduce the number of undetermined parameters. In addition, fewer parameters make it possible for this algorithm to be designed for data dimensionality reduction.

This paper selects design parameters that minimize the limit of misclassification rate and can be obtained in the corresponding closed form without using standard cross-validation methods, thus providing lower complexity and higher classification performance. Furthermore, the computational cost is also reduced compared to classic R-LDA in \cite{elkhalil2017asymptotic}. Through a comparative analysis of different data sets, our proposed classifier outperforms other popular classification techniques, such as improved LDA (I-LDA) in \cite{sifaou2020high}, support vector machine (SVM), $k$- Nearest neighbor (KNN), and a fully connected neural network (CNN).

The remainder of this article is organized as follows: Section \ref{mainresult} introduces the binary classification problem and the basic form of the SRLDA classifier. Section \ref{sec:3} presents a consistent estimation of the true error and a parameter optimization method. The SRLDA classifier with optimal intercept is improved in Section \ref{sec:4}. Section \ref{sec:5} extends the spectral correction regularization classification method to multi-classification problems. Section \ref{sec:6} compares the algorithm effects on simulated data and real data respectively. The final section analyzes the conclusion.

\section{Binary Classification and SRLDA Classifier}\label{mainresult}
{O}{n} the basis of the analysis framework in \cite{zollanvari2015generalized}, some modeling assumptions are restated and made throughout the paper. First, we have individually sampled binary classification sample sets $\fC_0$ and $\fC_1$ given in (\ref{eq:wlda}). Separate sampling is very common in biomedical applications, where data from two classes are collected without reference to the other, for instance, to discriminate two types of tumors or to distinguish a normal from a pathological phenotype.

A second assumption is that the classes possess a common covariance matrix. $\Pi_i$ follows a multivariate Gaussian distribution $N({\bmu}_i, {\bSig})$, for $i = 0, 1$, where ${\bSig}$ is nonsingular. Although it does not fully correspond to reality, the LDA still often performs better than quadratic discriminant analysis (QDA) in most cases, even with different covariances, because of the advantage of its estimation method \cite{raudys1980dimensionality}.

In this paper, we improve the LDA classifier in the particular scenarios, as the third assumption, wherein $\bSig$ takes the following particular form \cite{bai2012estimation}:
\begin{eqnarray}\label{pcm}
	\bSig&=&\sigma^2\left(\fI_p+\sum_{j\in \mathbb{I}}\lambda_j\fv_j\fv_j^T\),
\end{eqnarray}
where $\mathbb{I}=:\mathbb{I}_1 \cup \mathbb{I}_2$, $\mathbb{I}_1:= \{1,\dots, r_1\}$, $\mathbb{I}_2 := \{-r_2,\dots,-1\}$, $\sigma^2>0$, $r=r_1+r_2$, $\lambda_1\geq\cdots\geq\lambda_{r_1}>0>\lambda_{-r_2}
\geq\cdots\geq\lambda_{-1}>-1$ ($\lambda_{j}=\lambda_{p+j+1}$ for $j \in \bbI_2$), $\fI_p$ is a $p\times p$ identity matrix and $\fv_i$ ($i\in \mathbb{I}$) are orthogonal. This is the spiked model defined in this article.

The spiked model is often used to detect the number of signals embedded in noise \cite{nadakuditi2008sample, nadler2010nonparametric, L1986On, fazli2011?1}, or the number of factors in financial econometrics \cite{bai2002determining, onatski2009testing}, where the number of signals (or factors)  corresponds to the number of large spiked eigenvalues, while small spiked eigenvalues are often ignored. It should be pointed out that there are currently few research papers discussing the situation of small spiked eigenvalues. We believe that this is mainly due to two reasons. First, small spiked eigenvalues have no actual physical meaning in signal detection problems, so they are often ignored; second, there are not many pioneering contributions in the study of small spiked eigenvalues theory, which is not enough to arouse widespread interest among scholars.
However, in real data analysis, we found that small spiked eigenvalues can play a unique role in algorithm improvement. This is because small spiked eigenvalues are more hidden in the noise. By introducing the assumption of small spiked eigenvalues in (\ref{pcm}), the modeling effect can be greatly improved.

Under the high-dimensional random matrix theory framework, population parameters such as the number of spiked eigenvalues and the estimators of spiked eigenvalues, the main parameters of this model, have been extensively and in-depth studied \cite{bai2012estimation,  2020Estimating, ke2021estimation, jiang2021generalized, fazli2011?1}. These works have greatly improved the ability to estimate the population covariance matrix and adapt to different data backgrounds. For the sake of simplicity, we assume that $\sigma^2$, $r_1$, $r_2$,  and $\lambda_i$ ($i\in I$) are perfectly known. In our numerical simulations, we have used the method of \cite{bai2012estimation, jiang2021generalized, ke2021estimation} to estimate the unknown parameters in (\ref{pcm}).

In this paper, the spectrally-corrected method is to correct the spectral elements of the sample covariance matrix to those of $\Sigma$. Start the eigen decomposition of the pooled covariance matrix as
\beq\label{eq:l_j}
\fS = \sum_{j=1}^pl_j\fu_j\fu_j^T,
\eeq
with $l_j$ being the $j$-th largest eigenvalue and $\fu_j$ its corresponding eigenvector. Then correct $l_j$ to the corresponding one of $\Sigma$ as follows:
\begin{eqnarray}\label{eq:tfS}
	\tfS&=&\sigma^2\(\fI_p+\sum_{j\in \bbI}\lambda_{j}\fu_j\fu_j^T\),
\end{eqnarray}
where   $\fu_{j} \triangleq \fu_{p+1+j}$ for $j \in \mathbb{I}_2$. In (\ref{eq:tfS}), $\tfS$ is totally same as (\ref{pcm}) except the spiked eigenvectors. In this way, the original structure of $\Sigma$ is preserved as much as possible, but the disadvantage is that $\widetilde{\fS}$ is a biased estimator of $\Sigma$ because of the bias of sample spiked eigenvectors from that of the population. How to deal with the bias? R-LDA \cite{friedman2001elements, zollanvari2015generalized} method inspires us to introduce regularization parameters for the sample spiked matrix for LDA. Then we have the spectrally-corrected and regularized discriminant analysis (SRLDA) function, that is
\begin{eqnarray}\label{eq:wrlda}
	W^{SRLDA}\left(\overline{\fx}_0, \overline{\fx}_1, \tfH, \fx\right)=\sigma^{-2}\left(\fx-\frac{\overline{\fx}_0 + \overline{\fx}_1}{2}\right)^T\tfH\left(\overline{\fx}_0
	-\overline{\fx}_1\right),
\end{eqnarray}
where
\begin{eqnarray}\label{eq:tfH}
	\tfH&=&\left(\fI_p+\gamma_1\sum_{j\in\bbI_1}\lambda_{j}\fu_j\fu_j^T
	+\gamma_2\sum_{j\in\bbI_2}\lambda_{j}\fu_{j}\fu_{j}^T\right)^{-1}.
\end{eqnarray}
Here $\gamma_1$ and $\gamma_2$ are designed parameters to be optimal. We assume that
\begin{eqnarray}\label{eq:A}
	\(\gamma_1,\gamma_2\)\in\mathbb{A}=\left\{\gamma_1,\gamma_2:0<\gamma_1,\gamma_2<M, \gamma_2\notin\cup_{j\in\bbI_2}U\(|\lambda_{j}|^{-1},\delta\)\right\}
\end{eqnarray}
for some up bound $M$ and any small $\delta>0$, where $U\(|\lambda_{j}|^{-1},\delta\)=\left\{x: \left|x-|\lambda_{j}|^{-1}\right|<\delta\right\}$. From a practical point of view, we only need  to assume that $\gamma_2\notin\cup_{j\in\bbI_2}U\(|\lambda_{j}|^{-1},\delta\)$ to ensure that the spectral norm of $\tfH$ is bounded. However, the restriction to range $\mathbb{A}$ is needed later for the proof of the uniform convergence results. The designed SRLDA classifier is then given by
\begin{eqnarray}
	\psi_n^{SRLDA}=\left\{
	\begin{array}{ll}
		1, & \hbox{if $W^{SRLDA}\left(\overline{\fx}_0, \overline{\fx}_1, \tfH, \fx\right)\leq c$;} \\
		0, & \hbox{if  $W^{SRLDA}\left(\overline{\fx}_0, \overline{\fx}_1, \tfH, \fx\right)> c$,}
	\end{array}
	\right.
\end{eqnarray}
where $c=\log(\pi_1/\pi_0)$. Given $\overline{\fx}_0, \overline{\fx}_1, \tfS$, the error contributed by class $i$ ($=0$ or $1$), is defined by the probability of misclassification $\varepsilon_i^{SRLDA}$, that is 
\begin{eqnarray}\label{eq:Visrlda}
	&&P\left((-1)^iW^{SRLDA}\left(\overline{\fx}_0, \overline{\fx}_1, \tfH, \fx\right)\leq(-1)^ic\left|\fx\in\Pi_i,\overline{\fx}_0, \overline{\fx}_1, \tfH\right.\right)\nonumber\\
	\label{eq:Visrlda}
	&=&\Phi\left(\frac{(-1)^{i+1}G\left(\bmu_i,\overline{\fx}_0, \overline{\fx}_1, \tfH\right)+(-1)^i\sigma^2c}{\sqrt{D\left(\overline{\fx}_0, \overline{\fx}_1, \tfH, \bSig
			\right)}}\right),
\end{eqnarray}
and $\Phi(\cdot)$ denotes the cumulative distribution function of a standard normal random variable and
\begin{eqnarray}
	G\left(\bmu_i,\overline{\fx}_0, \overline{\fx}_1, \tfH\right)&=&\left(\bmu_i-\frac{\overline{\fx}_0+
		\overline{\fx}_1}{2}\right)^T\tfH\left(\overline{\fx}_0-
	\overline{\fx}_1\right),\\
	D\left(\overline{\fx}_0, \overline{\fx}_1, \tfH, \bSig\right)&=&\left(\overline{\fx}_0-
	\overline{\fx}_1\right)^T\tfH\bSig\tfH\left(\overline{\fx}_0-
	\overline{\fx}_1\right).
\end{eqnarray}
The true error of $\psi_n^{SRLDA}$ is expressed as:
\beq\label{eq:errorrate}
\varepsilon^{SRLDA}= \pi_0\varepsilon_0^{SRLDA} + \pi_1 \varepsilon_1^{SRLDA}.
\eeq

\section{Consistent Estimate of the True Error and Parameter Optimization}\label{sec:3}

{I}{n} this section, we derive a consistent estimator of the true error of SRLDA based on random matrix theory used to obtain optimal parameters. The theoretical optimal $\bgamma = \left(\gamma_1, \gamma_2\right)$ is chosen to minimize the total misclassification rate:
\beq\label{eq:min0}
\bgamma^*=\mathop{\arg \min}\limits_{\bgamma\in \mathbb{A}}\varepsilon^{SRLDA}(\bgamma),
\eeq
and its consistent estimate is responsible for the acquisition of optimal empirical parameters. Under the asymptotic growth regime, the following assumptions will help us to build a consistent estimate.

\begin{assumption}\label{as:1}
	$p, n_0, n_1\rightarrow \infty$, and the following limits exist: $\frac{p}{n_0}\rightarrow J_0>0$, $\frac{p}{n_1}\rightarrow J_1>0$ and $\frac{p}{n}\rightarrow J<1$ where $n=n_0+n_1$.
\end{assumption}

Assumption \ref{as:1} is the norm assumption in the framework of high-dimensional random matrix theory, under which this paper constructs the spectral-corrected method to correct the bias of the classical empirical estimation.

\begin{assumption}\label{as:2} For $J<1$,
	$r_1$ and $r_2$ are fixed and $\lambda_1>\cdots>\lambda_{r_1}>\sqrt{J}>0>-\sqrt{J}
	>\lambda_{-r_2}>\cdots>\lambda_{-1}>-1$, independently of $p$ and $n$.
\end{assumption}

The assumption \ref{as:2} is the basis of our analysis and the basic premise of applying high-dimensional random matrix theory, which guarantees a one-to-one mapping between sample eigenvalues and unknown population eigenvalues. In fact, when $\lambda_j>\sqrt{J}$ (or $-1<\lambda_j<-\sqrt{J}$), $\lambda_j$ can be consistently estimated for $j\in \mathbb{I}$.

\begin{assumption}\label{as:3}
	$\|\bmu_i\|$ has a bounded Euclidean norm, that is $\|\bmu_i\|=O(1)$, $i=0,1$.
\end{assumption}

\begin{assumption}\label{as:4}
	The spectral norm of $\Sigma$ is bounded, that is $\|\bSig\|=O(1)$.
\end{assumption}


\begin{theorem}\label{thm:1}
	Under Assumptions \ref{as:1} to \ref{as:4}, for any $\bgamma\in \bbA$  and $k=0$ (or $1$), we have
	\beq
	&\left|\bG_k(\bgamma)-\frac{(-1)^k}{2}\|\bmu\|^2\overline{G}
	(\bgamma)+\frac{\sigma^2}{2}
	\(\frac{p}{n_0}-\frac{p}{n_1}\)\right|\xrightarrow{a.s.} 0,\label{eq:3_2}\\
	&\left|\bD(\bgamma)
	-\sigma^2\|\bmu\|^2\overline{D}(\bgamma)-\sigma^4\(\frac{p}{n_0}+\frac{p}{n_1}\)\right| \xrightarrow{a.s.} 0,\label{eq:3_3}
	\eeq
	uniformly, where $\bG_k(\bgamma)=G\left(\bmu_k,\overline{\fx}_0, \overline{\fx}_1, \tfH\right)$, $\bD(\bgamma)=D\left(\overline{\fx}_0, \overline{\fx}_1, \tfH, \Sigma\right)$,
	\beq
	\overline{G}\left(\bgamma\right)
	&=&1-\sum_{i=1,2}\sum_{j\in \bbI}a_jb_j\gamma_{i,j}\delta_{i,j},\label{eq:oG}\\
	\overline{D}(\bgamma)
	&=&1+ \sum_{j\in \bbI}\lambda_jb_j
	-2\sum_{i=1,2}\sum_{j\in \bbI}\(\lambda_j+1\)a_jb_j\gamma_{i,j}\delta_{i,j}\nonumber\\
	&&+\sum_{i=1,2}\sum_{j\in \bbI}a_j b_j\(\lambda_ja_j+1\)\gamma_{i,j}^2\delta_{i,j}.\label{eq:oD}
	\eeq
	Here $\bmu = \bmu_0-\bmu_1$, $\gamma_{i,j}=\frac{\gamma_i\lambda_j}{1+\gamma_i\lambda_j}$,
	$a_j = \frac{\lambda_j^2-c}{\lambda_j(\lambda_j+c)}$, $b_j = \frac{\(\bmu^T\fv_j\)^2}{\|\bmu\|^2}$ for $i=1,2$, $j\in \bbI$
	and
	\beqq
	\delta_{i,j}=\left\{
	\begin{array}{ll}
		1, & \hbox{$i=1, j>0$ or $i=2, j<0$;} \\
		0, & \hbox{otherwise.}
	\end{array}
	\right.
	\eeqq
\end{theorem}

\begin{prop}
	See Appendix A.
\end{prop}

According to Theorem \ref{thm:1}, a deterministic equivalent of the global misclassification rate can be obtained as:
\beqq
\varepsilon^{SRLDA}(\bgamma)
-\overline{\varepsilon}^{SRLDA}(\bgamma)\xrightarrow{a.s.}0,
\eeqq
where $\overline{\varepsilon}^{SRLDA}(\bgamma)$ is defined as
\beq\label{eq:varepsilon}
\pi_0\Phi\(-\frac{\overline{G}(\bgamma)-\eta}
{2\sqrt{\alpha}\sqrt{\overline{D}(\bgamma)+\zeta}}\)
+\pi_1\Phi\(-\frac{\overline{G}(\bgamma)+\eta}
{2\sqrt{\alpha}\sqrt{\overline{D}(\bgamma)+\zeta}}\),
\eeq
and $\alpha= \frac{\sigma^2}{\|\bmu\|^2}$, $\eta= \alpha\(\frac{p}{n_0}-\frac{p}{n_1} +2c\)$,
$\zeta=\alpha\(\frac{p}{n_0}+\frac{p}{n_1}\)$, respectively. (\ref{eq:varepsilon}) is similar to (17) given in \cite{sifaou2020high}, but here are only two regularized parameters that are optimized on a more specific bounded set in Theorem \ref{thm:3}.

\begin{theorem}\label{thm:2}
	Under the setting of Assumptions \ref{as:1} to \ref{as:4}, we have
	\beq\label{eq:thm2}
	{\varepsilon}^{SRLDA}(\bgamma)
	-\overline{\varepsilon}^{SRLDA}(\bgamma)
	\xrightarrow{a.s.}0,\;\;\;\forall \bgamma\in\mathbb{A},
	\eeq
	uniformly, where $\overline{\varepsilon}^{SRLDA}$ and $\mathbb{A}$ are given in (\ref{eq:varepsilon}) and (\ref{eq:A}) respectively.
\end{theorem}

\begin{prop}According Theorem \ref{thm:1}, (\ref{eq:uGb}) and the continuity of $\Phi$, (\ref{eq:thm2}) holds naturally.
\end{prop}

\begin{theorem}\label{thm:3}
	Under the setting of Assumptions \ref{as:1} to \ref{as:4}, the optimal parameters $\bgamma=\left(\gamma_1^*,\gamma_2^*\right)$ that minimize $\overline{\varepsilon}^{SRLDA}(\bgamma)$ are given by:
	\beqq
	\gamma_1^* = \frac{\om_1^*}{(1-\om_1^*)\lambda_1} \;\;\mbox{and}\;\;\gamma_2^* = \frac{-\om_2^*}{(1-\om_2^*)\lambda_{-1}},
	\eeqq
	where $\bom^* = (\om_1^*,\om_2^*)$ is the minimizer of the function
	\beqq
	f(\bom)=
	\pi_0\Phi\(-\frac{1}{2\sqrt{\alpha}}\frac{\widetilde{G}
		(\bom)-\eta}{\sqrt{\widetilde{D}(\bom)+\zeta}}\)
	+\pi_1\Phi\(-\frac{1}{2\sqrt{\alpha}}\frac{\widetilde{G}
		(\bom)+\eta}{\sqrt{\widetilde{D}(\bom)+\zeta}}\),
	\eeqq
	and $\bom = \(\om_{1},\om_2\)\in{\bbB}$.
	Here $$\bbB=\left\{\bom = (\om_1,\om_2):0<\om_1,\om_2<1, \om_2\neq \(1+\lambda_{j}/\lambda_{-1}\)^{-1}, j\in \bbI_2\right\},$$
	\beqq
	\widetilde{G}\left(\bom\right)
	&=&1-\sum_{i=1,2}\sum_{j\in \bbI}a_jb_j\gamma_{i,j}\delta_{i,j},\\
	\widetilde{D}(\bom)
	&=&1+ \sum_{j\in \bbI}\lambda_jb_j
	-2\sum_{i=1,2}\sum_{j\in \bbI}\(\lambda_j+1\)a_jb_j\gamma_{i,j}\delta_{i,j}\\
	&&+\sum_{i=1,2}\sum_{j\in \mathbb{I} }a_jb_j\(\lambda_ja_j+1\)\gamma_{i,j}^2\delta_{i,j},\\
	\gamma_{1,j}&=&\frac{\om_1\lambda_j}{(1-\om_1)\lambda_1+\om_1\lambda_j},
	\;\;\;j\in\bbI_1,\\
	\gamma_{2,j}&=&\frac{-\om_2\lambda_{j}}{(1-\om_2)
		\lambda_{-1}-\om_2\lambda_{j}},
	\;\;\;j\in\bbI_2.
	\eeqq
	And $\eta$ and $\zeta$ are defined in Theorem \ref{thm:1}.
\end{theorem}

\begin{prop}
	See Appendix B.
\end{prop}

\begin{theorem}\label{th:lrmt}
	Under Assumptions \ref{as:1} to \ref{as:4}, for any deterministic unit vector $\fw\in S_{\mathbb{R}}^{p-1}$, we have
	\begin{equation}\label{eq:angle}
		\left\langle \fw, \fu_j\fu_j^T\fw \right\rangle \xrightarrow{a.s.} \dfrac{\lambda_j^2-J}{\lambda_j(\lambda_j+J)}\left\langle \fw, \fv_j\fv_j^T\fw \right\rangle ,
	\end{equation}
	for $j\in \bbI$.
\end{theorem}


\section{Improved SRLDA with optimal intercept}\label{sec:4}
{T}{he} intercept part of (\ref{eq:wlda}) is another important parameter that affects the misclassification rate. Since the empirical estimation of the intercept part always causes severe bias in unbalanced classes and high-dimensional settings, it is necessary to consider bias correction, which is now a general procedure to minimize the misclassification rate such as \cite{chan2009scale,huang2010bias,wang2018dimension}. In this section, we apply the bias correction procedure to the improved LDA classifier and name the classifier ``OI-SRLDA'' to refer to the optimal-intercept-SRLDA classifier.p Starting off from our proposed classifier and replacing the constant term with a parameter $\theta$, the score function can be written as:

\begin{eqnarray*}\label{eq:worlda}
	W^{OI-SRLDA}\left(\overline{\fx}_0, \overline{\fx}_1, \tfH, \fx\right)=\sigma^{-2}\left(\fx-\frac{\overline{\fx}_0 +\overline{\fx}_1}{2} \right)^T \tfH\left(\overline{\fx}_0-\overline{\fx}_1\right)+\sigma^2\theta,
\end{eqnarray*}
where
\begin{eqnarray*}
	\tfH&=&\left(\fI_p+\gamma_1\sum_{j=1}^{r_1}\lambda_{j}\fu_j\fu_j^T+
	\gamma_2\sum_{j=1}^{r_2}\lambda_{-j}\fu_{-j}\fu_{-j}^T\right)^{-1},
\end{eqnarray*}
and $ \theta $ is a parameter that will be optimized. The corresponding misclassification rate $\varepsilon^{OI-SRLDA}(\bgamma,\theta)$ is expressed as:

\begin{eqnarray*}
	\pi_0 \Phi\left(\frac{-\bG_0(\bgamma)
		-\sigma^2\theta } {\sqrt{\bD(\bgamma)}}\right)
	+ \pi_1 \Phi\left(\frac{\bG_1(\bgamma) +\sigma^2\theta } {\sqrt{\bD(\bgamma)}}\right).
\end{eqnarray*}
The asymptotic equivalent of $\varepsilon^{OI-SRLDA} $ can be obtained by following similar steps as in previous section, that is:
\beqq
\left|\varepsilon^{OI-SRLDA}(\bgamma,\theta)
-\overline{\varepsilon}^{OI-SRLDA}\left(\bgamma,\theta\right)\right|\xrightarrow{a.s.}0,
\eeqq
where $\overline{\varepsilon}^{OI-SRLDA}(\bgamma,\theta)$ is given as:
\begin{eqnarray*}
	\pi_0\Phi\(\frac{-\overline{G}(\bgamma)
		+\eta(\theta)}{2\sqrt{\alpha}\sqrt{\overline{D} (\bgamma)+\zeta}}\)
	+\pi_1\Phi\(\frac{-\overline{G}(\bgamma)
		-\eta(\theta)}{2\sqrt{\alpha}\sqrt{\overline{D}(\bgamma)+\zeta}}\).
\end{eqnarray*}
in which $\eta(\theta)=\alpha\left(\frac{p}{n_0}-\frac{p}{n_1} -2\theta\right)$.
By standard optimization solving methods, we have the optimal parameter $ \theta^* $ as following:
$$ \theta^*=\frac{1}{2} \left(\frac{p}{n_0}-\frac{p}{n_1} \right) -\frac{\overline{D} (\bgamma)+\zeta }{\overline{G}(\bgamma)} \log\frac{\pi_1}{\pi_0}. $$
Replacing $ \theta^* $ by its expression, the asymptotic misclassification rate becomes $$\overline{\varepsilon}^{OI-SRLDA}(\bgamma)=
\pi_0\Phi\(\overline{\Delta}_1+\overline{\Delta}_2\)
+\pi_1\Phi\(\overline{\Delta}_1-\overline{\Delta}_2\), $$ in which $\overline{\Delta}_1(\bgamma)=
\frac{-\overline{G}(\bgamma)}{2\sqrt{\alpha}
	\sqrt{\overline{D}(\bgamma)+
		\zeta}}$ and $\overline{\Delta}_2(\bgamma)=\frac{\sqrt{\alpha} \sqrt{\overline{D}(\bgamma) +\zeta}}{\overline{G}(\bgamma)} \log\frac{\pi_1}{\pi_0}$.
Then we are able to find the new optimal parameter vector $\bgamma = (\gamma_1,\gamma_2) $ that minimizes the asymptotic
misclassification rate $ \overline{\varepsilon}^{OI-SRLDA}$.

\begin{theorem}\label{thm:5}
	The optimal parameters $\bgamma^* =(\gamma_1^*,\gamma_2^*)$ that minimize {{$\overline{\varepsilon}^{OI-SRLDA}$}} are given by:
	\beq\label{eq:varepsilon3}
	\gamma_1^* = \frac{\om_1^*}{(1-\om_1^*)\lambda_1}, \;\;\mbox{and}\;\;\gamma_2^* = \frac{\om_2^*}{(1-\om_2^*)\lambda_{-1}},
	\eeq
	where $\bom^*=(\om_1^*,\om_2^*)$ is the minimizer of the function $$f(\bom)= \pi_0\Phi\(\widetilde{\Delta}_1+
	\widetilde{\Delta}_2\)+\pi_1\Phi\(\widetilde{\Delta}_1
	-\widetilde{\Delta}_2\),$$ and
	$\widetilde{\Delta}_1(\bom)=\frac{-\widetilde{G}(\bom)}
	{2\sqrt{\alpha}\sqrt{\widetilde{D}(\bom)+\zeta}}$,
	$\widetilde{\Delta}_2(\bom) = \frac{\sqrt{\alpha} \sqrt{\widetilde{D}(\bom) +\zeta}}{\widetilde{G}(\bom)} \log\frac{\pi_1}{\pi_0}$, respectively.
	Here
	$\widetilde{G}\left(\bom\right)$, $\widetilde{D}(\bom)$ and $\zeta$ are defined in Theorem \ref{thm:3}.
\end{theorem}

\begin{theorem}\label{thm:6}
	Under Assumptions \ref{as:1} to \ref{as:4}, we have
	$$\left| \frac{1}{\alpha}-\frac{1}{\hat{\alpha}}\right|\stackrel{a.s,} {\longrightarrow}{0},\quad \left|\lambda_j-\hat{\lambda}_j\right|\stackrel{a.s,} {\longrightarrow}{0},\quad
	\left|b_j-\hat{b}_j\right|\stackrel{a.s,} {\longrightarrow}{0},$$
	where,  for any $j\in\bbI$,
	\beq \label{hat_alpha}
	\frac{1}{\hat{\alpha}}&=&\frac{\left\| \hat{\bmu}\right\|^2}{\sigma^2}-J_1-J_0,
	\eeq
	\beq \label{hat_lammda}
	\hat{\lambda}_j=\frac{1}{2}\(l_j-1-J+\mbox{sgn}(J)\sqrt{\(l_j+1-J\)^2-4l_j}\),
	\eeq
	\beq \label{hat_bj}
	\hat{b}_j=\frac{1+J/\hat{\lambda}_j}{1-J/\hat{\lambda}_j}
	\frac{\hat{\bmu}^T\fu_j\fu_j^T\hat{\bmu}}
	{\|\hat{\bmu}\|^2-J_1\sigma^2-J_0\sigma^2},
	\eeq
	$\hat{\bmu}=\overline{\fx}_0- \overline{\fx}_1$, $l_j$ is given in (\ref{eq:l_j}), and $sgn(\cdot)$ is a sign function.
\end{theorem}

\begin{prop} The proof is a direct application of the equation (2.7) in Bai and Ding, 2012\cite{bai2012estimation} and it's thus omitted here.
\end{prop}

\section{The Spectrally-corrected and Regulized Technique for fisher's discriminant in the case of multiple classes}\label{sec:5}

Suppose we have $K$ distinct classes and samples from each class follow a $p$-dimensional multivariate normal distribution with mean vector $\bmu_k$ and covariance matrix $\Sigma$ with the form (\ref{pcm}), where $k = 0,1,2,...,K-1$. Assume we observe $n_k$ i.i.d. random samples from the $k$th class, that is,
\beq
\Pi_k:\;\;\bx_{1,k},\bx_{2,k},\cdots,\bx_{n_k,k}
\stackrel{\mathrm{i.i.d.}}{\sim}N(\bmu_k,\bSig)
\eeq
The total sample size is then $n=\sum_{i=0}^{k-1}n_k$. Let $\pi_k$ denote the prior probability of observing sample from the $k$th class with $\sum_{k=0}^{K-1}\pi_k=1$.

According the generalization of Fisher's discriminant for multiple classes, and the assumption that the dimensionality $p$ is greater than the number $K$, there is a subspace with  $(K-1)$ dimensionality at most, to maximize the between-class distance and minimize the within-class one. That is to obtain the optimal projection matrix $\fW^*$ to maximize
\beq\label{eq:multW}
J\(\fW^*\) = \max_{\fW} \mbox{Tr}\left\{\(\fW\bSig\fW^T\)^{-1}\(\fW\bSig_b\fW^T\)\right\}
\eeq
in which
\beqq
\bSig_b = \sum_{k = 0}^{K-1}\pi_k\(\bmu_k-\bmu\)\(\bmu_k-\bmu\)^T,
\eeqq
with $\bmu = \sum_{k=0}^{K-1}\pi_k\bmu_k$. The optimal solution $\fW^*$ of (\ref{eq:multW}) is determined by those eigenvectors of $\bSig^{-1}\bSig_b$ that correspond to the $K$ largest eigenvalues.

By the spectrally-corrected technique, the estimation of $\bSig$ is taken as the same form as $\tfH^{-1}$ (\ref{eq:tfH}) in which $\bu_j$s are the corresponding eigenvectors of the sample-based within-classes scalar matrix $\fS_w$,
\beq\label{eq:Sw}
\fS_w = \frac{1}{n}\sum_{k=0}^{K-1}\sum_{j=1}^{n_k}\(\bx_{j,k}-\overline{\bx}_k\)\(\bx_{j,k}-\overline{\bx}_k\)^T
\eeq
and $\bSig_b$ is estimated by the between-classes scatter matrix $\fS_b$,
\beqq
\fS_b = \sum_{k=0}^{K-1}\frac{n_k}{n}\(\overline{\bx}_k-\overline{\bx}\)\(\overline{\bx}_k-\overline{\bx}\)^T
\eeqq
in which $\overline{\bx}$ is the pooled sample mean. The parameters $\gamma_1$ and $\gamma_2$ can be determined by cross validation or minimizing the following the joint error rate function
\beqq
\mbox{Error}= \sum_{{i=0}}^{K-1}\pi_i\sum_{{j=0}\atop{j\neq i}}^{K-1} \varepsilon_{i,j}^{SRLDA}
\eeqq
in which $\varepsilon_{i,j}^{SRLDA}$ is the true error with the form (\ref{eq:Visrlda}) where $\Pi_0$ and $\Pi_1$ are $\Pi_i$ and $\Pi_j$ respectively and $\fS_w$ in (\ref{eq:Sw}) is instead of the within-class scatter matrix of $\Pi_i$ and $\Pi_j$.

\section{Simulations and experiments}\label{sec:6}
{I}{n} this part, we first simulate and verify that the SRLDA classifier outperforms those of the classical LDA, RLDA and ILDA under the spiked model. Then, we applied actual data to analyze the good performance of SRLDA in classification and dimensionality reduction. In the section \ref{subsec:1},	for the handwritten digit data set, the accuracy rate of SRLDA, RLDA, ILDA, CLDA, SVM, KNN, and CNET classifiers was comparatively analyzed. In the section \ref{subsec:2}, under the kernel transform and the different principle component analysis reduction, the SRLDA classifier performs better than that in the before section. In the section \ref{subsec:3}, we provide the good performance of SRLDA in the dimensionality reduction. 

\subsection{Simulated data}

In this part, we use the following Monte Carlo method to estimate the true corrected rate:
\begin{itemize}
	\item Step 1: Set $\sigma^2 = 1$ and choose $r_1=3$, $r_2 = 1$ orthogonal directions $\fv_1$, $\fv_2$, $\fv_3$, $\fv_p$ with respect to the corresponding weights $\lambda_1 = 20$, $\lambda_2 = 10$, $\lambda_3 = 5$, $\lambda_p = -0.99$. Let $\mu_0=\frac{a}{\sqrt{p}}(1,\cdots,1)^{'}$, $\mu_1=-\mu_0$.
	\item Step 2: Generate $n_i(m_i)$ training samples (testing samples) for class $i=0$, $1$ in which $\pi_0 = \frac{n_0}{n_0+n_1} = \frac{m_0}{m_0+m_1}$ and is given as $0.1$, $0.2$, $0.3$, $0.4$, $0.5$.
	\item Step 3: Using training sample, derive the corrected spectral of the sample covariance and the optimal parameter $\gamma^{*}_1$ and $\gamma^{*}_2$ of the SRLDA method as discribed in the section \ref{sec:3} using grid search over $\{w_1,w_2\}\in{\{[0,1)\times{[0,1)}\}}$ with adjustable accuracy.
	\item Step 4: Determine the optimal parameter $\gamma^*$ of RLDA by grid search over $\gamma\in\{10^{i/10}, i=-10:1:10\}$, by using the training set.
	\item Step 5: Esimate the accouracy rates of LDA, classical LDA, RLDA, I-LDA and SRLDA methods using a set of 10000 testing samples.
	\item Step 6: Repeat Step 2-4, 500 times and determine the average classification accuracy (or error) rate of these classifiers.
\end{itemize}

\begin{figure}
	\setlength{\abovecaptionskip}{0pt}
	\setlength{\belowcaptionskip}{0pt}
	\centering
	\subfloat[\label{fig:comparea} Scatter points between eigenvalues of $\fS_w$ and $\fS_{mp}$]{
		\includegraphics[height=4.11cm,width=9cm]{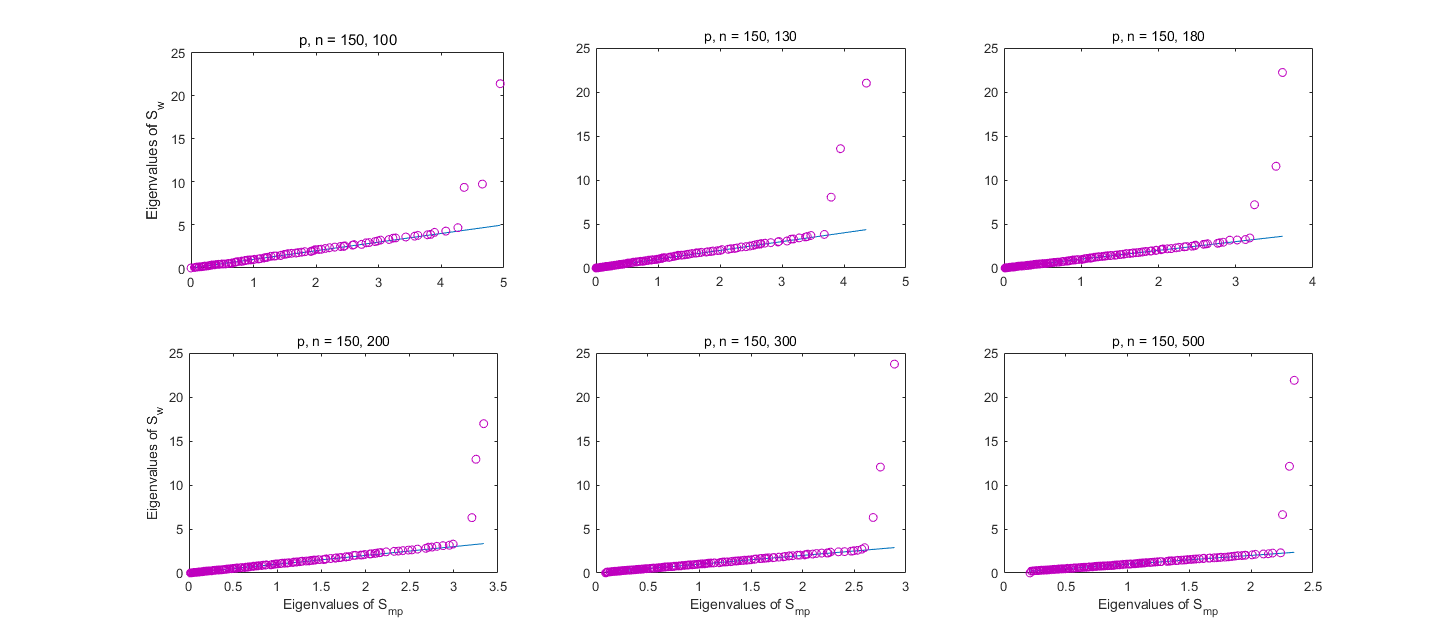}}\\%
	\subfloat[\label{fig:compareb} Scatter points between inverses of eigenvalues $\fS_w$ and $\fS_{mp}$]{
		\includegraphics[height=4.11cm,width=9cm]{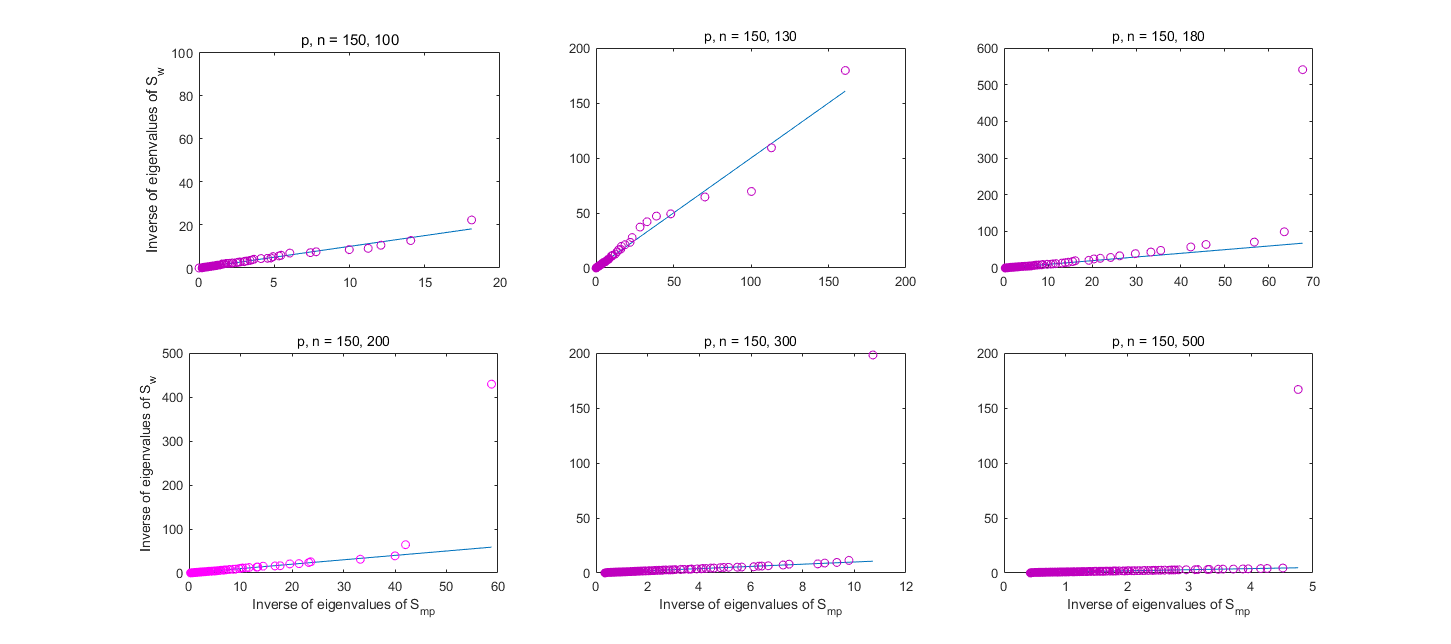}}
	\caption{Scatter points between (inverses of) eigenvalues of $\fS_w$ and $\fS_{mp}$, where $\fS_{mp}$ is generated by the same way as $\fS_w$ with $r_1=r_2=0$.}
	\label{fig:compare}
\end{figure}

\begin{figure}
	\setlength{\abovecaptionskip}{0pt}
	\setlength{\belowcaptionskip}{0pt}
	\centering
	\subfloat[\label{fig:a} TLDA, RLDA, ILDA and SCRLDA]{
		\includegraphics[height=4.11cm,width=8.56cm]{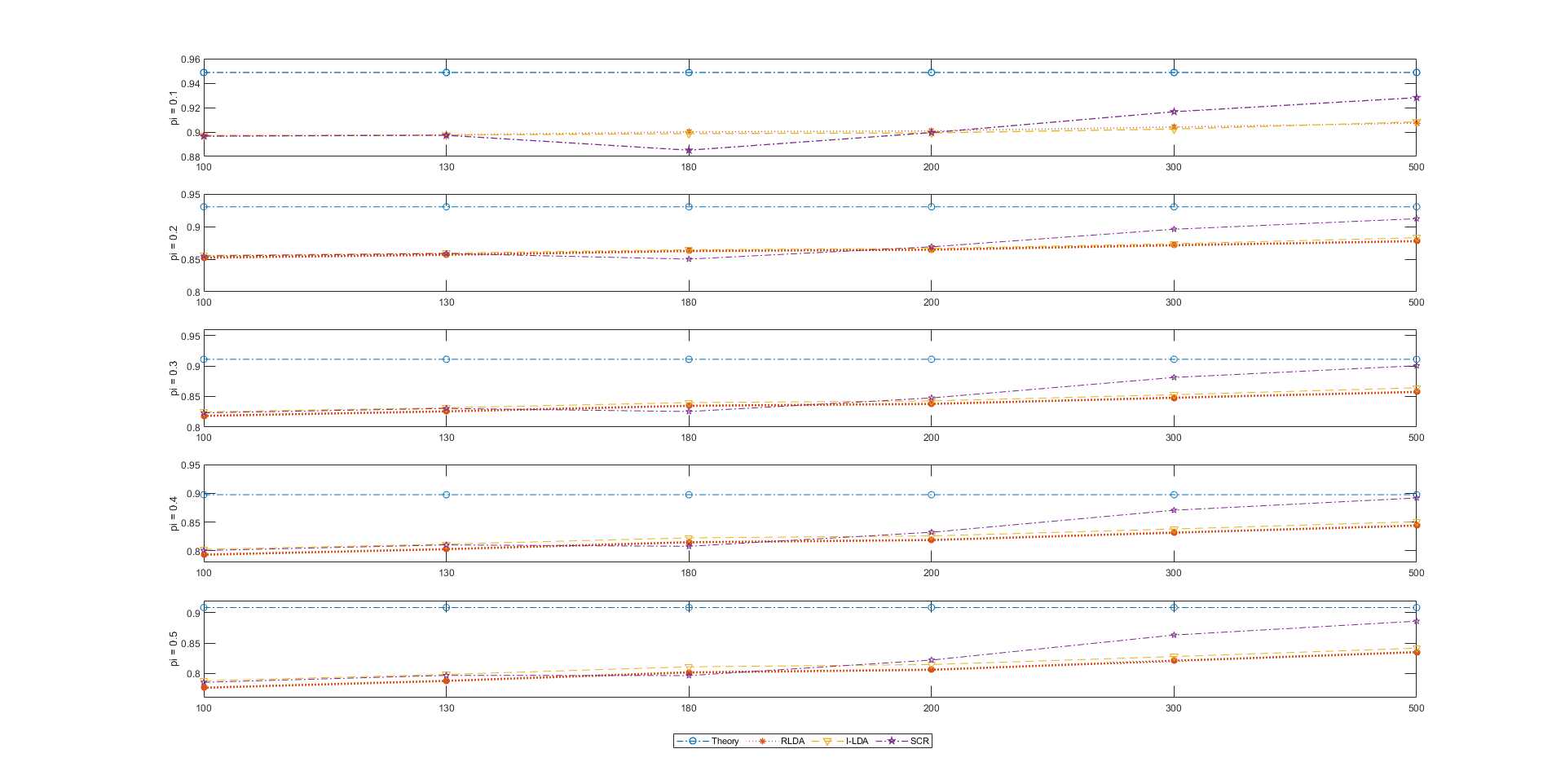}}\\%
	\subfloat[\label{fig:b} Classical LDA]{
		\includegraphics[height=3.11cm,width=7.56cm]{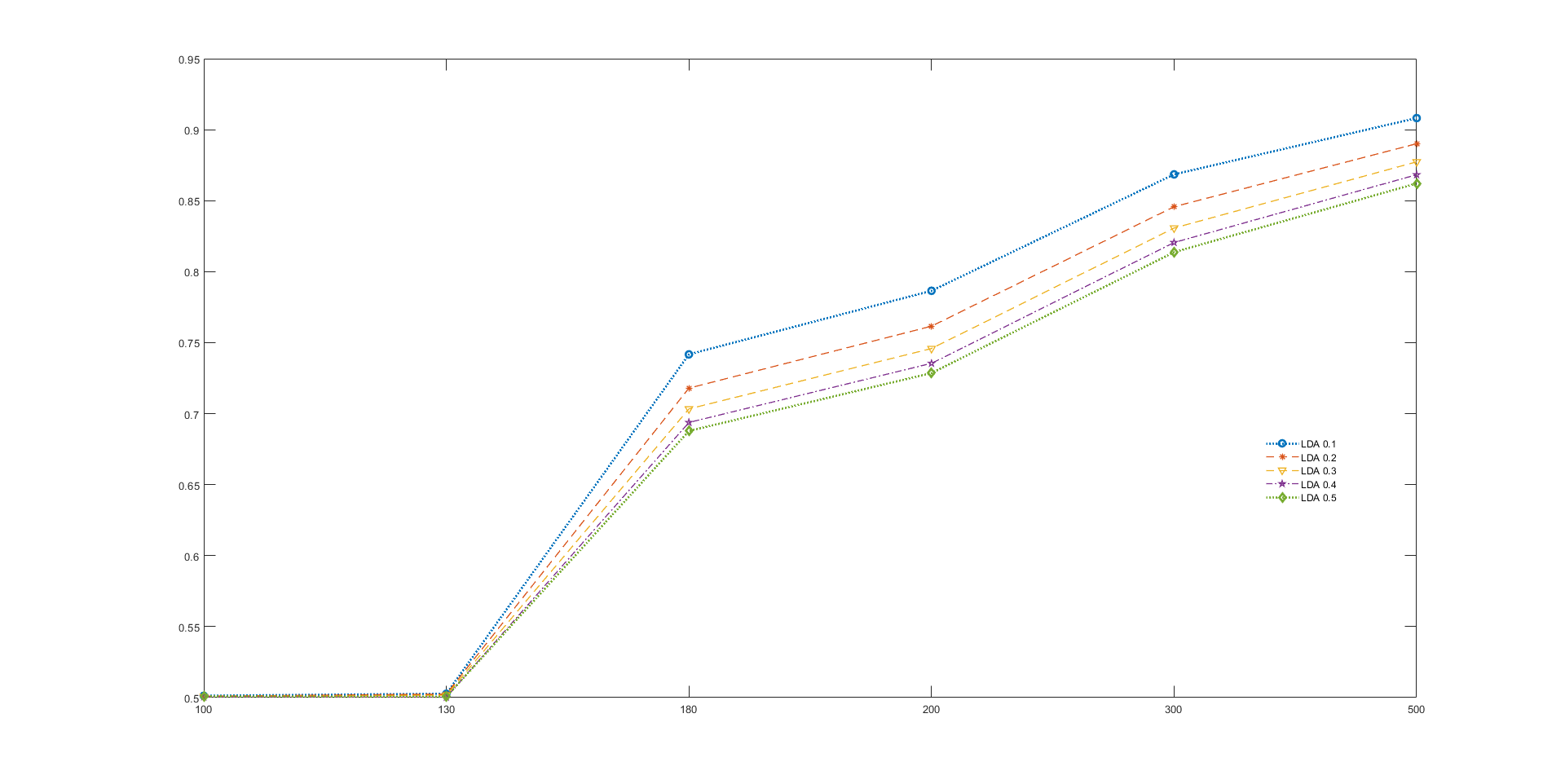}}
	\caption{Corrected rate vs. sample size $n$ for $p=150$ and $\pi_0=0.1, 0.2, 0.3, 0.4, 0.5$. Comparison for LDA, RLDA, ILDA and SRLDA with  simulated data in which TLDA is the theoretical linear discriminant classifier.}
	\label{fig1}
\end{figure}

From Step 1 and Step 2, given $r_1 = r_2 = 0$, we can generate the sample set without the spiked structure, that $\Sigma = \fI_p$ and the corresponding pooled sample covariance, noted as $\fS_{mp}$. Figure \ref{fig:comparea} shows the scatter points of the eigenvalues of $\fS_w$ v.s. $\fS_{mp}$ with the sample size $n = 100, 130, 180, 200, 300, 500$ for $n = 150$. The large spiked eigenvalues, designed in Step 1, can be detected from the eigenvalues of $\fS_w$ for all $n$ cases. For the small spiked eigenvalues, designed in Step 1, it is difficult to detect when $n<p$. But as $n>p$, inverses of eigenvalues of $\fS_w$ show the information of the small spiked eigenvalues clearly in Figure \ref{fig:compareb}. In Figure \ref{fig1}, we provide the average classification accuracy rates  of LDA, classical LDA,  RLDA, I-LDA and SRLDA classifiers with the simulated data, in which  sample size $n=100,130,180,200,300,500$ for $p=150$ and $\pi_0=0.1,0.2,0.3,0.4,0.5$. When $n<p$, let $r_2 = 0$ and $J = 1$.

\subsection{Real Data}

For empirical analysis, we use the ``MNIST Database of Handwritten Digits'' in \href{https://archive.ics.uci.edu/dataset}
{UCI Machine Learning Repository.\footnote{https://archive.ics.uci.edu/dataset/683/mnist+database+of+handwritten+digits}} In the well-known database, there are 70,000 handwritten digits (10 class labels) with each example represented as an image of 28 x 28 gray-scale pixels.

\subsubsection{The case of two classes}\label{subsec:1}
In this part, we uses the following protocol for the real dataset:
\begin{itemize}
	\item Step 1: Select sample sets of two classes  from ``MNIST Database of Handwritten Digits'' as $\fC_0$ and $\fC_1$. Denote $n$ as the number of training samples in which $n_0$ and $n_1$ are the numbers of samples of $\fC_0$ and $\fC_1$, respectively, and satisfied the ratios $\frac{n_0}{n}$ and $\frac{n_1}{n}$. The remaining samples are used as a test dataset in order to estimate the true accuracy rate.
	
	\item Step 2: Select the dimensions with sample covariance zero and select the train dateset with sample sizes $(500, 1000, 1500, 2000, 2500)$.
	
	\item Step 3: Using the training dataset, design the SRLDA, RLDA and I-LDA by the same way given in Section \ref{sec:5}, and SCLDA is SRLDA with $\gamma_1=\gamma_2 = 1$.
	
	\item Step 4: Compute the classifier by SVM, KNN and CNET. For SVM, all the hyper-parameters optimized one is selected. For KNN, we select the optimal knn number in the set $\{1,3,5,...,49\}$, that is, odd numbers from one to fifty. 
	
	\item Step 4: Using the test dateset, estimate the true accuracy for classifiers.
	
	\item Repeat steps 1-4 500 times and determine the average accuracy of both classes.

\end{itemize}
\begin{figure}
	\setlength{\abovecaptionskip}{0pt}
	\setlength{\belowcaptionskip}{0pt}
	\centering
	\includegraphics[height=4.11cm,width=8.56cm]{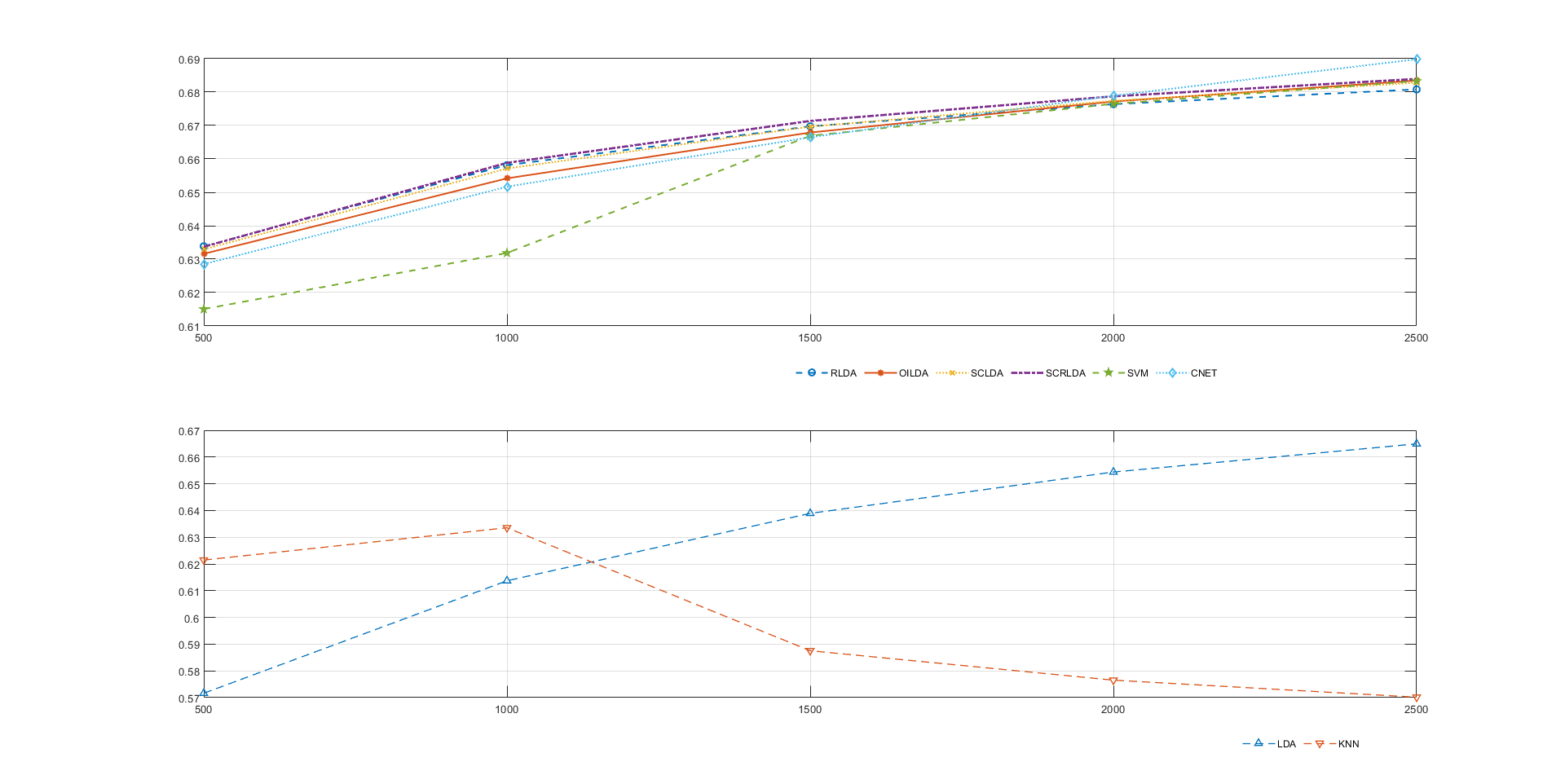}
	\caption{Comparison between the proposed classifier, RLDA, I-LDA, SCLDA, SRLDA, SVM, CNET, KNN and LDA using MNIST Database of Handwriten Digits $3$ and $8$.}
	\label{fig2}
\end{figure}

\subsubsection{The classification of two classes by kernel transformation technology and PCA reduction}\label{subsec:2}
\begin{figure}
	\setlength{\abovecaptionskip}{0pt}
	\setlength{\belowcaptionskip}{0pt}
	\centering
	\includegraphics[height=4.11cm,width=8.56cm]{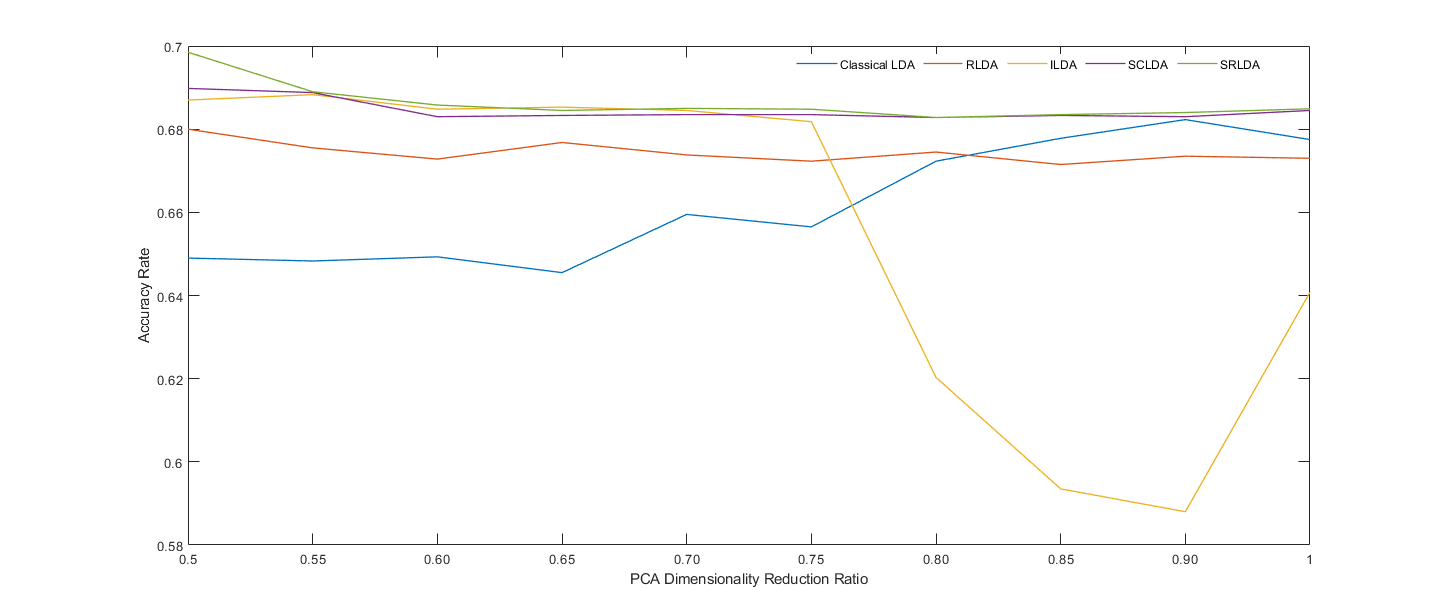}
	\caption{Comparison between the proposed classifier, RLDA, I-LDA, SCLDA, SRLDA and Classical LDA using MNIST Database of Handwritten Digits $3$ and $8$ with PCA.}
	\label{fig:pca}
\end{figure}

\begin{figure}
	\setlength{\abovecaptionskip}{0pt}
	\setlength{\belowcaptionskip}{0pt}
	\centering
	\includegraphics[height=4.11cm,width=8.56cm]{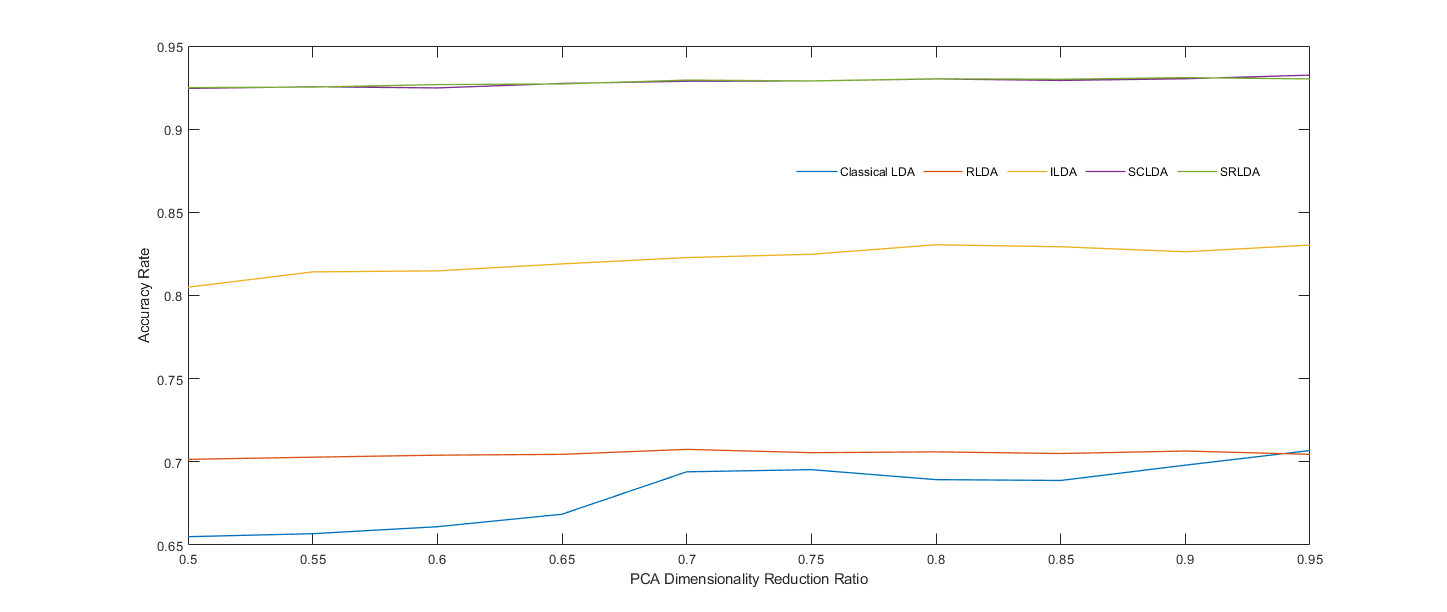}
	\caption{Comparison between the proposed classifier, RLDA, I-LDA, SCLDA, SRLDA and Classical LDA using MNIST Database of Handwritten Digits $3$ and $8$ with Kernel transformation and PCA.}
	\label{fig:kernelpca}
\end{figure}

This section compares and analyzes the superior performance of the SRLDA algorithm after data kernel transformation (or not) and principal component analysis (PCA) dimensionality reduction. Five different kernel transformation methods are selected, such as linear kernel, polynomial kernel, Gaussian kernel, Sigmoid kernel, and Laplacian kernel, and the best-corrected rate is provided. Considering the problem of calculation speed, here we choose a data set of handwritten digits 3 and 8 with a sample size of 4000 and a proportion of 0.5. Figures \ref{fig:pca} and \ref{fig:kernelpca} show that the correction rates of SRLDA and SCLDA are better than Classical LDA, RLDA and ILDA. In particular, under kernel transformation, the advantages of SRLDA and SCLDA are more significant. 

\begin{figure}
	\setlength{\abovecaptionskip}{0pt}
	\setlength{\belowcaptionskip}{0pt}
	\centering
	\includegraphics[height=4.11cm,width=8.56cm]{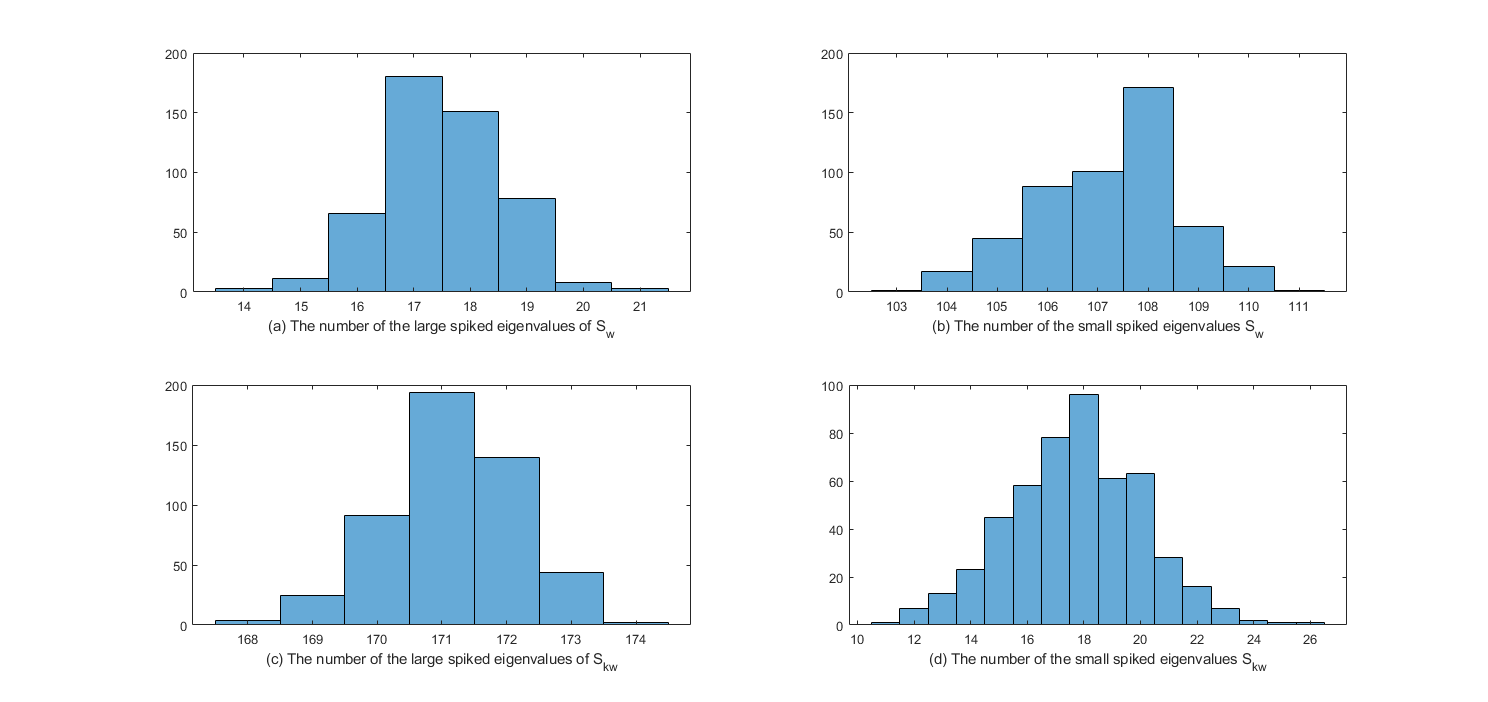}
	\caption{The histogram of the number of the spiked eigenvalues, where $\fS_w$ and $\fS_{kw}$ are the pooled sample covariance of the samples and the kernel transformed samples, respectively.}
	\label{fig:number_spiked}
\end{figure}

To further explain the working principle of the SC (SR) method, we show the distribution of the number on the large spiked eigenvalues and the small eigenvalues when the PCA dimensionality reduction ratio is 0.9 in Figure \ref{fig:number_spiked}. From the histogram, we can easily conclude that the number of small spiked eigenvalues is so large that it cannot be ignored.

\subsubsection{The case of multiple classes}\label{subsec:3}

\begin{figure}
	\setlength{\abovecaptionskip}{0pt}
	\setlength{\belowcaptionskip}{0pt}
	\centering
	\includegraphics[height=4.11cm,width=8.56cm]{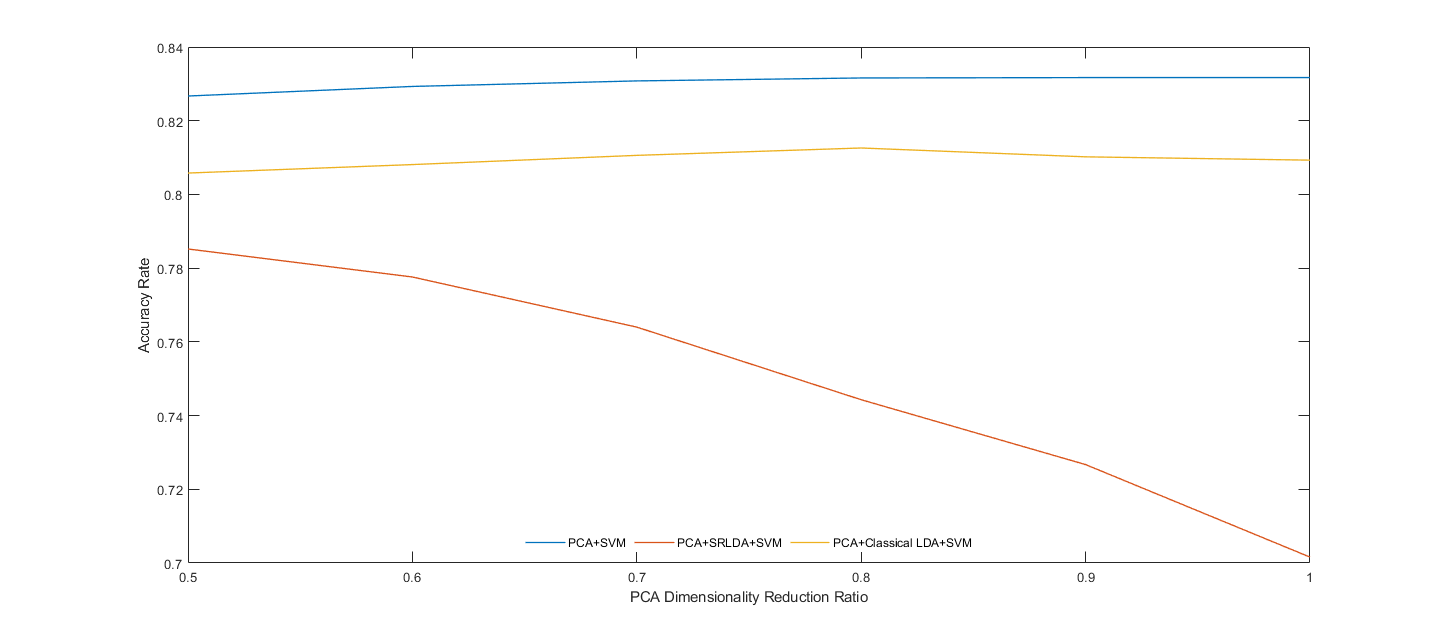}
	\caption{Comparison of the dimensionality reduction effects of SRLDA and classic LDA for face recognition data of different dimensions under PCA.
	}
	\label{fig:multi_classes}
\end{figure}

This section applies the SRLDA method to feature selection and extraction in multi-classification problems. Here we choose ``the Labeled Faces in the Wild (LFW) people dataset'', with sample size 1288, 450 features, and 7 classes, to test the dimensionality reduction effects of SRLDA, LDA under PCA.  In Figure \ref{fig:multi_classes}, under the same dimensionality conditions, classic LDA and SRLDA reduce the dimension of the data from more than 225 dimensions to only six dimensions but the loss in accuracy is not high, about 0.019 to 0.13. In particular, the accuracy of the SRLDA algorithm after dimensionality reduction is significantly higher than that of classic LDA, and this advantage becomes more obvious as the dimension increases.

\section{Conclusion} \label{sec:7}{
	
	{U}{sing}  real and simulated data sets for effective comparison, the SCRDA method has the following design points and excellent performance.
	\begin{itemize}
		\item[(1)] Retain the original model design structure as much as possible, that is, the spike covariance model. In this way, the results of eigenvalue estimation in high-dimensional settings and the limiting properties of the corresponding eigenvectors are fully utilized.
		\item[(2)] The regularization method once again proved its important role in reducing the risk (misclassification rate).
		\item[(3)] Based on empirical analysis, the SRLDA method is quite competitive with SVM, KNN, and CNN, but its algorithm is more lightweight and faster in the calculation.
		\item[(4)] This algorithm is also suitable for multi-classification problems and can be used for supervised dimensionality reduction. The dimensionality reduction effect is significantly better than traditional dimensionality reduction methods.
	\end{itemize}
	
	As a further extension, the same idea can be extended to other classification methods, such as quadratic discriminant classifiers or other spike covariance models.

	\bibliographystyle{plain}
	\bibliography{mainbib}

\end{document}